\documentclass[conference]{IEEEtran}
\IEEEoverridecommandlockouts
\usepackage{cite}
\usepackage{amsmath,amssymb,amsfonts}
\usepackage{algorithmic}
\usepackage{graphicx}
\usepackage{textcomp}
\usepackage{xcolor}
\usepackage{hyperref}
\usepackage{pifont}

\newcommand{\cmark}{\ding{51}}%
\newcommand{\xmark}{\ding{55}}%

\def\BibTeX{{\rm B\kern-.05em{\sc i\kern-.025em b}\kern-.08em
    T\kern-.1667em\lower.7ex\hbox{E}\kern-.125emX}}
\begin{document}

\title{Cross-Stitched Multi-task Dual Recursive Networks for Unified Single Image Deraining and Desnowing\\
\thanks{This research has been supported by the European Commission within the context of the project RESCUER, funded under EU H2020 Grant Agreement~101021836.}
}

\author{\IEEEauthorblockN{Sotiris Karavarsamis}
\IEEEauthorblockA{\textit{ITI} - \textit{CERTH}\\
Thessaloniki, Greece \\
skaravarsamis@iti.gr}
\and
\IEEEauthorblockN{Alexandros Doumanoglou}
\IEEEauthorblockA{\textit{ITI} - \textit{CERTH}\\
Thessaloniki, Greece \\
aldoum@iti.gr}
\and
\IEEEauthorblockN{Konstantinos Konstantoudakis}
\IEEEauthorblockA{\textit{ITI} - \textit{CERTH}\\
Thessaloniki, Greece \\
k.konstantoudakis@iti.gr}
\and
\IEEEauthorblockN{Dimitrios Zarpalas}
\IEEEauthorblockA{\textit{ITI} - \textit{CERTH}\\
Thessaloniki, Greece \\
zarpalas@iti.gr}
}

\maketitle

\begin{abstract}
We present the Cross-stitched Multi-task Unified Dual Recursive Network (CMUDRN) model targeting the task of unified deraining and desnowing in a multi-task learning setting. This unified model borrows from the basic Dual Recursive Network (DRN) architecture developed by Cai et al. The proposed model makes use of cross-stitch units that enable multi-task learning across two separate DRN models, each tasked for single image deraining and desnowing, respectively. By fixing cross-stitch units at several layers of basic task-specific DRN networks, we perform multi-task learning over the two separate DRN models. To enable blind image restoration, on top of these structures we employ a simple neural fusion scheme which merges the output of each DRN. The separate task-specific DRN models and the fusion scheme are simultaneously trained by enforcing local and global supervision. Local supervision is applied on the two DRN submodules, and global supervision is applied on the data fusion submodule of the proposed model. Consequently, we both enable feature sharing across task-specific DRN models and control the image restoration behavior of the DRN submodules. An ablation study shows the strength of the hypothesized CMUDRN model, and experiments indicate that its performance is comparable or better than baseline DRN models on the single image deraining and desnowing tasks. Moreover, CMUDRN enables blind image restoration for the two underlying image restoration tasks, by unifying task-specific image restoration pipelines via a naive parametric fusion scheme. The CMUDRN implementation is available at \url{https://github.com/VCL3D/CMUDRN}.
\end{abstract}


\section{Introduction}
\label{sec:intro}
Rain and snow are two common weather conditions that naturally degrade imaging and the performance of intelligent applications, such as object detection or surveillance operations. Operationally, these weather conditions are detrimental to first responder missions at disaster sites, such as collapsed buildings by earthquakes, or search-and-rescue operations in mountain areas, in which human vision can be degraded severely. Therefore, technical means for human vision augmentation are important for restoring the capability of human vision. Towards this end, vision augmentation image restoration techniques for rain and snow are called to remove the natural visual artifacts in images for more clear vision, hence enabling finer operation in downstream computer vision (or other) tasks.

According to Sun, Ang Jr and Rus \cite{sun2019convolutional}, unified models that solve the single image deraining task usually remove the rain streaks from images, but they are unable to remove the effect of mist that is caused by rain. Similar problems are exhibited by methods that solve the single image desnowing task either in a unified model or in a single image restoration model. Another common drawback is that unified models or single image restoration methods around these problems are very slow at handling very large image inputs. The latter drawback is often linked to the large number of model parameters.

In this work, we contribute with a network architecture that targets the unified deraining and desnowing tasks that is simple, lightweight and fast. We call the proposed network ``\underline{C}ross-stitched \underline{M}ulti-task \underline{U}nified \underline{D}ual \underline{R}ecursive \underline{N}etwork`` (CMUDRN). The model is trained using basic, task-specific, recursive, convolutional feature transformations. This model can be trained to blindly alleviate the natural visual artifacts caused by rain and snow, using only a single deep neural network model. Our work is on par with recent efforts on the topic, such as the recent work by Chen et al. \cite{Chen_2022_CVPR}. Our experiments suggest that the proposed CMUDRN model performs comparably well with baseline models and can attain a small space and time footprint depending on the input size and the value of a model hyperparameter.

\section{Related work on unified models}
\label{sec:relatedwork}

\begin{figure*}[t]
     \centering
     \includegraphics[scale=0.3]{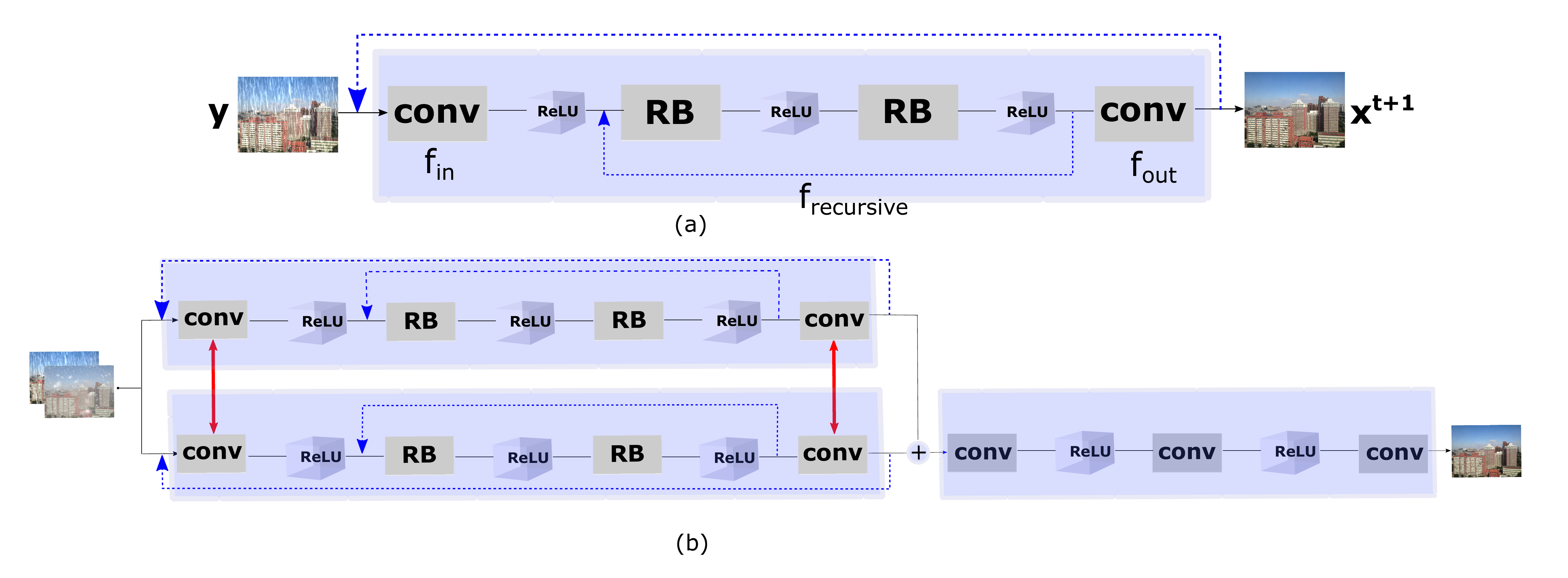}
     \caption{Illustration of (a) the basic Dual Recursive Network (DRN) developed by Cai et al. \cite{cai2019dual}, originally tasked for single image deraining; and, (b) the proposed Cross-stitched Multi-task Unified Dual Recursive Network (CMUDRN) that employs cross-stitch enabled multi-task learning of two separate DRN models, each tasked for single image deraining and desnowing, respectively. Feature map sharing is enabled via cross-stitch units due to Misra et al. \cite{misra2016cross}. Additionally, local supervision at the local, separate DRN models and the global, top fusion module regulate the local and overall function of the proposed CMUDRN model.}
     \label{fig:drn_based_models}
\end{figure*}

The related work on unified (also called ``all-in-one") image restoration around rain and snow has flourished in the last years, although the count of these works is much fewer than the studies which target these problems alone.

Chen et al. \cite{Chen_2022_CVPR} follow a teacher-student learning paradigm to learn student sub-networks that  specialize in denoising images from a particular weather type. The authors targeted a model for the unified handling of the deraining, desnowing and dehazing problems. Li, Tan and Cheong \cite{li2020all} proposed a unified model for rain, fog and snow, and for raindrop removal. The model has a generator and a discriminator submodule. The generator submodule has three feature extraction branches (one for each bad weather degradation), and feature search is used for feature encoding. Finally, a discriminator module is applied to classify the type of image degradation and clean the input image.

Li et al. \cite{AirNet} propose the all-in-one AirNet model targeting the task of blind image denoising, deraining and dehazing. The model uses contrastive learning to extract a neural representation of a particular bad weather condition or noise appearing in a given image, and then uses a similar representation learned off from training data in order to compute a clean image.

Sun, Ang Jr and Rus \cite{sun2019convolutional} propose a convolutional neural network-based model which is both fast to evaluate on data and also can solve single image deraining and dehazing caused by the veiling effect of rain streaks using a single deep neural network model employing the global information contained in images. The authors evaluate their real time-enabled model in the autonomous driving task, where scene segmentation and object detection is naturally degraded by the contamination of images by rain streaks.

The previously mentioned unified models \cite{Chen_2022_CVPR, li2020all, AirNet, sun2019convolutional}, even though they exhibit state-of-art performance on the evaluated datasets, their architecture and training procedure could be considered rather complex compared to the network that we propose in the present work. Moreover, their run-time performance is often neglected in the evaluation while our work focuses on the design of a lightweight and fast model for the unified denoising task. Last, \cite{sun2019convolutional} is one among a few works that can run in real-time. However, their proposed model is specifically engineered towards removing rain streaks and haze inside rainy images. Contrariwise, the proposed model does not use any weather-specific priors and aims to restore images from two different weather conditions, as opposed to the single condition of \cite{sun2019convolutional}.

\section{Multi-task learning}
\label{sec:multitasklearninng}

Multi-task learning was originally introduced in the work of Caruana \cite{caruanamulti}. A recent survey paper summarizing multi-task learning schemes designed specifically for deep learning models is due to Vandenhende et al. \cite{vandenhende2020revisiting}.

Here we build a two-branch deep CNN model via multi-task learning by using cross-stitch units, developed in prior work by Misra et al. \cite{misra2016cross}. Figure~\ref{fig:drn_based_models}(b) graphically illustrates the design of the multi-task learning-based model, while Figure~\ref{fig:drn_based_models}(a) shows the design of the DRN model developed by Cai et al. \cite{cai2019dual}. The latter model enables single-task deraining and desnowing, or essentially other related image restoration tasks such as dehazing. Cross stitch units were developed by Misra et al. \cite{misra2016cross}; they compute a linear combination of two feature maps, allowing to share features among convolutional feature maps from task-specific models. By doing so, a multi-task learning technique can enable the design of a more powerful model than in the setting where separate models are combined together without such a mechanism.

A cross-stitch unit is a feature map $\mathcal{W}$ (essentially, a tensor) that is a convex combination of two feature maps $\mathcal{W_{A}}$ and $\mathcal{W_{B}}$, obeying to the equation

\begin{equation}
\mathcal{W} = \alpha_{S} \mathcal{W_{A}} + (1 - \alpha_{S}) \mathcal{W_{B}}
\label{eq:cross_stitch_unit}
\end{equation}

By performing line search over the parameter $\alpha_{S}$, a better model can be identified in comparison to a model that combines separate single-task models. Throughout all the experiments reported in this paper, we set the parameter $a_{S}$ to $0.5$. Equation~\ref{eq:cross_stitch_unit} can be used to induce feature sharing among more than two feature maps by repetitively applying cross-stitch units on pairs of feature maps.

\section{Cross-stitched DRN models}
\label{sec:CDRN_model}

The DRN model due to Cai et al. \cite{cai2019dual} has a very simple architecture, and features a deep convolutional transformation with inter-locality and intra-locality feedback loops for two successive residual blocks \cite{he2016deep}. Empirically, these two types of loop create more powerful transformation mappings compared to no feedback loops at all, but they impact the time complexity of the model as more loops are required in the model design.

The basic model comprises a convolutional layer followed by a ReLU activation function, followed by a loop of two residual blocks (see He et al. \cite{he2016deep}) and then a final convolution layer followed by ReLU. An outer loop is added to the model that induces a feedback loop that joins the first layer down to the bottom layer, implying the recurrence relation

\begin{equation}
\mathbf{x^{t+1}} = \mathbf{y} + f_{out}(f_{recursive}(f_{in}(\mathbf{x^{t}},\mathbf{y}))), \ 1 \leq t \leq T
\label{eq:recurrence_loop}
\end{equation}
In the above equation, $\mathbf{y}$ is an input 3D tensor modelling an input RGB image. Essentially, $\mathbf{x^{1}} = \mathbf{y}$ is the initial condition of the recursive function, and $T$ is the maximum index of a term in the recurrence relation. Observing Figure~\ref{fig:drn_based_models}(a), $f_{in}$ is matched with the bottom convolutional feature map of the basic DRN model, $f_{recursive}$ matches with the middle double recursive block-unit loop, and $f_{out}$ is the top convolutional feature map.

Although DRN was initially developed for the single image deraining task, in this paper we suggest that the model is also capable of solving the desnowing task (see the preliminary baseline DRN model results on our augmented CSD dataset \cite{chen2021all} on Table~\ref{tab:baseline_drn_models}), achieving usable image restoration models with good peak signal-to-noise ratio (PSNR) and structural similarity (SSIM) values. An ablation study for the proposed CMUDRN model suggests that the herein proposed CMUDRN model hypothesis attains more increased or comparable performance to a number of essential baseline DRN models.

\section{Training the CMUDRN joint model}
\label{sec:training}
\begin{figure}[t]
     \centering
     \includegraphics[width=3.3in]{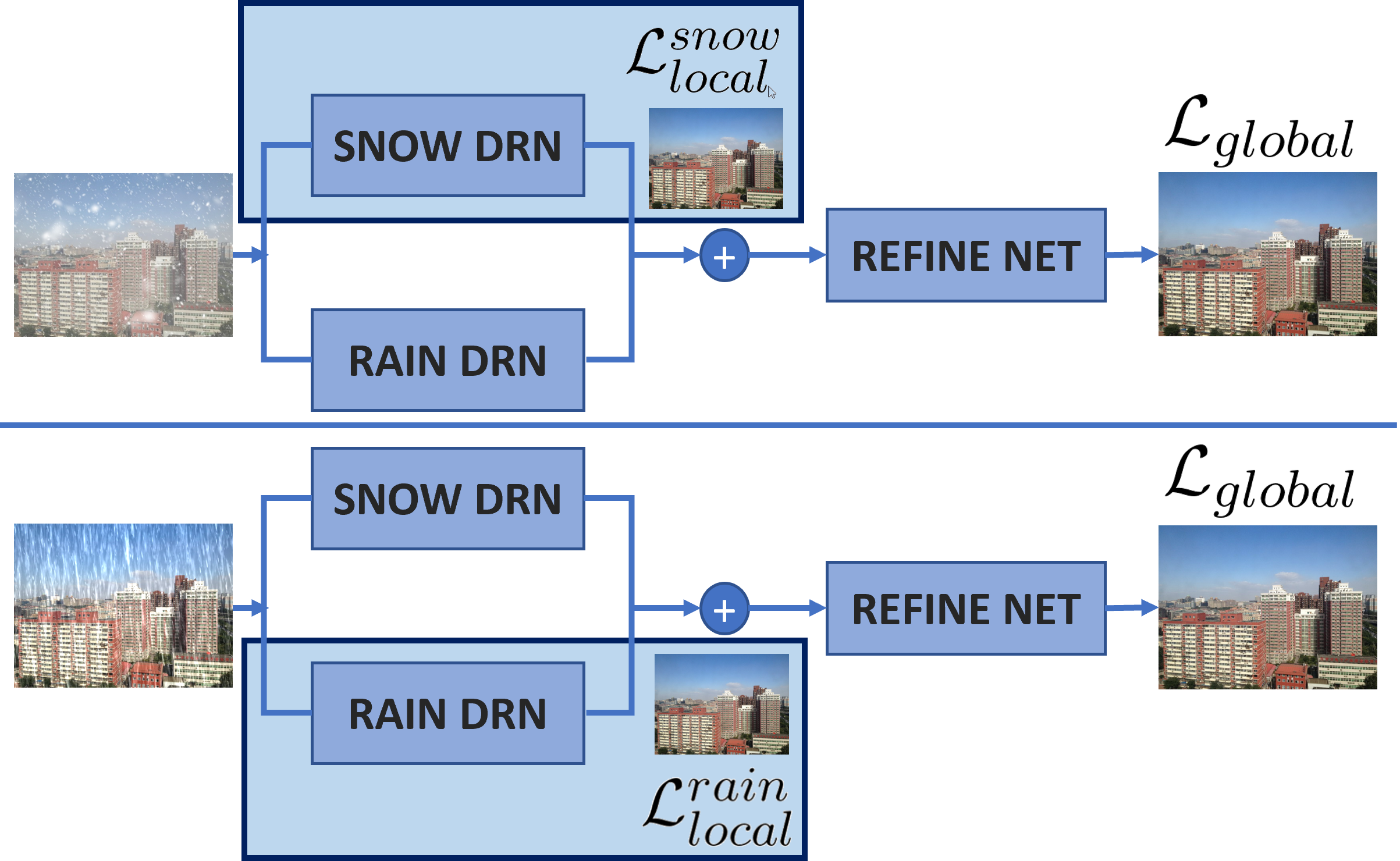}
     \caption{The training scheme of the proposed method involves different local loss functions applied to the output of the different DRN modules based on the weather conditions of the input image.}
     \label{fig:drn_train}
\end{figure}

All unified single image restoration models (such as those by Chen et al. \cite{Chen_2022_CVPR}; Chen et al. \cite{chen2021all}; or others), require $k$-tuples of degraded images from an initial image $\mathbf{x}$ (also denoted as $\mathbf{y}$ in Equation~\ref{eq:recurrence_loop}) to train the unified model. In order to train the proposed CMUDRN network, we made a design choice: we created a synthetic dataset with rain image examples, and snow images originally offered by the Comprehensive Snow Dataset (CSD) dataset by Chen et al. \cite{chen2021all}. The CSD dataset has $8,000$ snowy-scene images with corresponding groundtruth. To generate synthetic rain images for the groundtruth images, we used a MATLAB implementation\footnote{Implementation available at https://github.com/liruoteng/RainStreakGen} of the photorealistic rain generation algorithm by Garg and Nayar \cite{garg2006photorealistic}. Snowy images for the groundtruth images are already provided by the CSD dataset; therefore, we opted to reuse these available images and not to generate snowy images on our own.

To train the CMUDRN model, we begin by a given clean reference image $\mathbf{x}$, and we construct a tuple $(\mathbf{x_{r}},\mathbf{x_{s}})$, by augmenting the image as necessary for the respective weather condition. Subsequently, we pass the two impaired images into the network, each one in a different pass. At each pass, we supervise the DRN subnetwork that we dedicate for each weather condition with $\mathcal{L}_{local}^{*}$, based on the weather conditions of the input image. Thus, in each pass, a different DRN sub-network is supervised with \textit{local} supervision. In both passes though, the output of the refine-network (or fusion network) is supervised with the global supervision $\mathcal{L}_{global}$. Local supervision enforces each DRN submodule to specialize in solving the image restoration task for a specific weather condition. With the cross-stitch units though (Section \ref{sec:multitasklearninng}), the two DRN modules co-operate in a multi-task setting. Each of the two modules is gaining additional information from the experience learned from the opposite module while solving a similar image restoration task, although under different weather conditions. In this way, the proposed approach is able to tackle the unified image restoration task, while at the same time improving denoising performance by treating the unified problem in a multi-task setting. Fig. \ref{fig:drn_train} depicts the various components of the method.

\section{Local and global loss functions}
\label{sec:local_and_global_loss_functions}

For the design of the CMUDRN unified single image deraining and desnowing model, we utilize local loss functions and a global loss function to regulate learning. Let $\mathcal{X} = \{ (\mathbf{x_{i}^{R}}, \mathbf{x_{i}^{S}}) \ | \ i \in \{1,...,n\} \}$ be a training set of rain images $\mathbf{x_{i}^{R}}$ and snowy images $\mathbf{x_{i}^{S}}$, assuming a total of $n$ training tuples. The rain-specific DRN submodule is modelled as a tensor function $f_{R}(\mathbf{x}; \mathcal{W_{R}})$. Accordingly, $f_{S}(\mathbf{x}; \mathcal{W_{S}})$ is the tensor function for snow. Let also $f_{fusion}(\mathbf{x_{R}},\mathbf{x_{S}})$ be the fusion function that combines together the outputs $f_{R}(\mathbf{x}; \mathcal{W_{R}})$ and $f_{S}(\mathbf{x}; \mathcal{W_{S}})$ of the rain-specific and snow-specific DRN submodules.

The CMUDRN model optimizes a combined loss function $\mathcal{L}_{combined}$ that is the sum of a local loss function $\mathcal{L}_{local}$, a loss function counting loss values for tensors generated by the recursions in the DRN submodules, and a global loss function $\mathcal{L}_{global}$; namely

\begin{equation}
    \mathcal{L}_{combined} = \mathcal{L}_{local} +
    \mathcal{L}_{recur} +
    \mathcal{L}_{global}
    \label{eq:combined_loss_function}
\end{equation}
The $\mathcal{L}_{local}$ loss function is the sum of the loss functions $\mathcal{L}_{rain}$ and $\mathcal{L}_{snow}$, each of them being denoted as $\mathcal{L}_{*}$

\begin{equation}
    \sum_{i=1}^{n} \ell_{ssim}(\mathbf{x_{i}^{*}}, \mathbf{x_{i}^{GT}}) + ||  f_{*}(\mathbf{x_{i}^{*}}; \mathcal{W_{*}}) - \mathbf{x_{i}^{GT}}||_{F}^{2}
    \label{eq:loss_local}
\end{equation}
where $\ell_{ssim}(\mathbf{x},\mathbf{y}) = 1 - SSIM(\mathbf{x},\mathbf{y})$ is the loss function of structural dissimilarity and $\mathcal{L}_{local}$ is defined as

\begin{equation}
    \mathcal{L}_{local} = \mathcal{L}^{rain}_{local} + \mathcal{L}^{snow}_{local}
    \label{eq:loss_local_sum}
\end{equation}
The global loss function $\mathcal{L}_{global}$ regulates the function of the DRN subnetworks implicitly by considering the function of the fusion module. $\mathcal{L}_{global}$ is defined as

\begin{equation}
    \mathcal{L}_{global} = \sum_{i=1}^{n} \ell_{ssim}(\mathbf{\tilde{y}_{i}}, \mathbf{x_{i}^{GT}}) + \ell_{F}(\mathbf{\tilde{y}_{i}}, \mathbf{x_{i}^{GT}})
    \label{eq:loss_global}
\end{equation}
where $\mathbf{\tilde{y}_{i}} = f_{fusion}(\mathbf{x_{i}^{R}},\mathbf{x_{i}^{S}})$ and $\ell_{F}(\cdot,\cdot)$ is the Frobenius tensor norm loss function defined as $\ell_{F}(\mathbf{x},\mathbf{y}) = ||\mathbf{x} - \mathbf{y}||_{F}$.

The $\mathcal{L}_{recur}$ counts the loss value regarding the discrepancy among a ground-truth image $\mathbf{x_{i}^{GT}}$ and the estimated denoised image $\mathbf{x_{i}^{j}}$ at a recursive iteration $j$ for $j = 1, ..., T$. $T$ is the integer number of inter-loops and intra-loops that the model performs. We assume that inter-loop and intra-loop counts are equal in the CMUDRN model. We count $\mathcal{L}_{recur}$ for both the rain component as $\mathcal{L}_{recur}^{rain}$ and the snow component $\mathcal{L}_{recur}^{snow}$. Hence,

\begin{equation}
    \mathcal{L}_{recur} = \mathcal{L}_{recur}^{rain} + \mathcal{L}_{recur}^{snow}
    \label{eq:recurrence_loss}
\end{equation}
$\mathcal{L}_{recur}^{*}$ is defined for either rain or snow-related variables as

\begin{equation}
    \mathcal{L}_{recur}^{*} = \sum_{i=1}^{n} \sum_{r=1}^{T} \ell_{ssim}(\mathbf{x_{i,r}^{*}},\mathbf{x_{i}^{GT}}) + ||f_{*}(\mathbf{x_{i,r}^{*}}) - \mathbf{x_{i}^{GT}}||_{F}^{2}
    \label{eq:recurrence_loss_perproblem}
\end{equation}
where $\mathbf{x_{i,r}^{*}}$ is the estimated dependent variable of the recurrence Equation~\ref{eq:recurrence_loop} at iteration $1 \leq r \leq T$ for the rain or snow problem.

Finally, the function $f_{fusion}(\cdot,\cdot)$ implementing the fusion module of the CMUDRN network is given by

\begin{equation}
        f_{fusion}(\mathbf{x_{i}^{R}},\mathbf{x_{i}^{S}}; \mathcal{W_{R}}, \mathcal{W_{S}}, \mathcal{W_{F}}) = g(\mathbf{x_{i}^{R}} + \mathbf{x_{i}^{S}})
    \label{eq:fusion_module}
\end{equation}
where $g(\cdot)$ is the convolutional transformation
\begin{equation}
    g(\mathbf{x}) = Conv_{3}(ReLU(Conv_{2}(ReLU(Conv_{1}(\mathbf{x})))))
    \label{eq:fusion_module_conv_transformation}
\end{equation}
$Conv_{1}$ is a $6 \times 16$ feature map; $Conv_{2}$ is a $16 \times 16$ feature map; and, $Conv_{3}$ is a $16 \times 3$ feature map. In function $g(\cdot)$, the independent variable $\mathbf{x}$ is set to be $\mathbf{x} = \mathbf{\tilde{x}_{1}} \boxplus \mathbf{\tilde{x}_{2}}$ (where $\boxplus$ is the tensor concatenation operation across the channels dimension). Since $\mathbf{\tilde{x}_{1}}$ and $\mathbf{\tilde{x}_{2}}$ are $m \times 3 \times W \times H$ tensors, then $\mathbf{x}$ is a $m \times 6 \times W \times H$ tensor.

\section{ablation study}
\label{sec:ablativestudy}

\begin{figure*}
\centering
\includegraphics[scale=0.45]{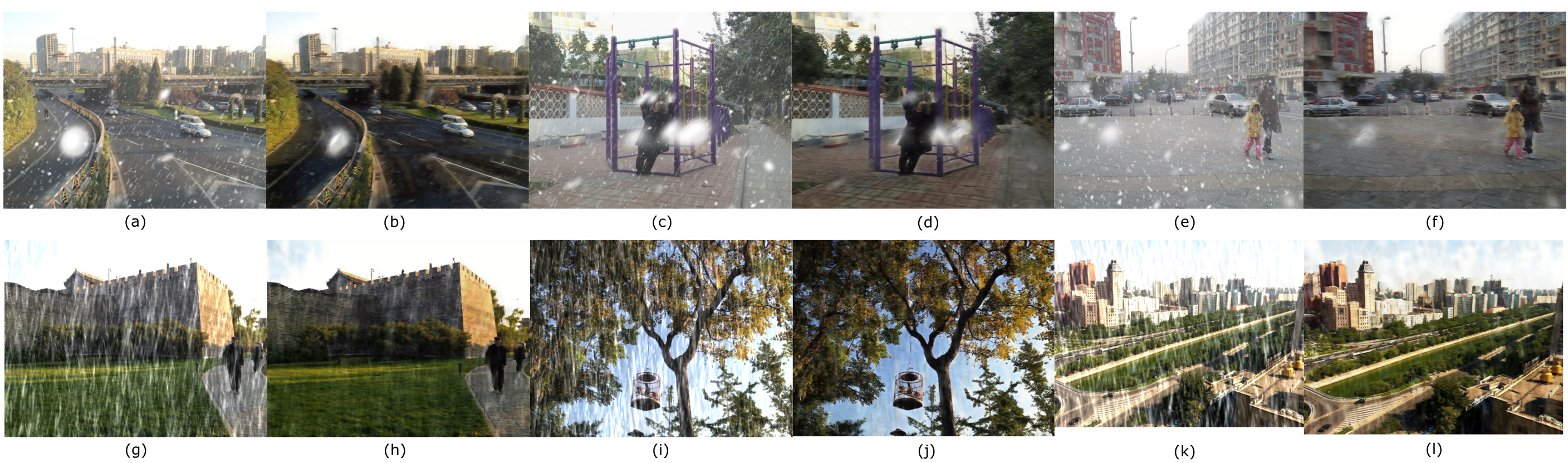}
\caption{Example pairs of a snowy image and the corresponding desnowed image from the CSD dataset (a)-(f), and corresponding examples from the augmented rain-image portion of the CSD dataset (g)-(l). Images in odd columns are noisy images, and those in even columns are the corresponding restored images by the proposed model.}
\label{fig:example_restored_images}
\end{figure*}

\begin{table}[]
\caption{Ablation experiments on the proposed CMUDRN model. We measure the impact of cross-stitch units in the model design; and the impact of loss functions for local and global supervised learning. We use the augmented CSD dataset as a testbed for these ablation experiments. In \textbf{black} we signify the best result; in \textcolor{blue}{\textbf{blue}}, we signify the second best result. $3$ intra and inter loops are used in each ablation experiment to obtain lightweight models. A higher loop count can empirically guide towards a better model.}
\label{tab:drn_model_ablation_study}
\centering
\begin{tabular}{ccc|cc|ll}
\multicolumn{3}{c}{\textbf{network components}} & \textbf{global} & \textbf{local} & \textbf{rain} & \textbf{snow}\\
CS & SSIM & Frob & \textbf{loss} & \textbf{losses} & PSNR/SSIM & PSNR/SSIM \\ \hline
\cmark & \cmark & \cmark & \cmark & \cmark    & \textbf{25.82} / 0.82 & \textbf{23.62} / 0.84\\
\xmark & \cmark & \cmark & \cmark  & \cmark & \textcolor{blue}{\textbf{25.21}} / 0.81 & \textcolor{blue}{\textbf{23.32}} / 0.82\\
\cmark & \xmark & \cmark & \cmark &  \cmark   & 25.07 / 0.80 & 23.11 / 0.82\\
\xmark & \xmark & \cmark & \cmark & \cmark  & 25.07 / 0.80 & 23.03 / 0.82\\
\cmark & \cmark & \xmark & \cmark &  \cmark   & 24.73 / \textbf{0.86} & 22.27 / \textbf{0.87}\\
\xmark & \cmark & \xmark &  \cmark & \cmark & 24.11 / \textcolor{blue}{\textbf{0.84}} & 22.32 / \textcolor{blue}{\textbf{0.87}}\\
\cmark & \cmark & \cmark & \cmark & \xmark & 24.69 / 0.81 & 22.97 / 0.83\\
\end{tabular}
\end{table}
We conduct an ablation study on the CMUDRN model using the observed PSNR and SSIM values as driver quantitative performance scores for model selection. Three components in the CMUDRN model are ablated in combinations and the empirical average PSNR and SSIM performance scores are observed. The components are: (a) the cross-stitching units allowing for the CMUDRN model to train correlated DRN submodules capable of single image deraining and desnowing; (b) the structural disimilarity (SSIM) loss function; and, (c) the Frobenius tensor norm loss function.

Table~\ref{tab:baseline_drn_models} renders six component ablation combinations for the CMUDRN model where one or more components in the CMUDRN are either used, or are ablated from the model.
The first six hypotheses consider ablated CMUDRN models using both local loss functions (each corresponding to a separate DRN submodule), and a global loss function that regulates the function of the CMUDRN model head; see Figure~\ref{fig:drn_based_models}(b). The last model hypothesis at the bottom of Table~\ref{tab:drn_model_ablation_study} serves as a baseline model hypothesis versus the first model hypothesis at the top of the same table, where local loss functions are applied during the training of the model to supervise the input-output behavior of the separate DRN submodules (each being adapted to restoring rainy and snowy images, respectively). This last experiment was conducted to monitor the function of the CMUDRN model without local loss functions. In this case, the local DRN models are trained without local supervision and only the loss function of the network head is applied during training.

We draw the following conclusions by observing the performance scores of the six CMUDRN model hypotheses: (a) the first model hypothesis that combines applied cross-stitch units, and SSIM and Frobenius tensor norm loss functions attains the best empirical performance on the last $30\%$ chunk of the CSD dataset being reserved for model testing. With this hypothesis, we observe that the model attains a better performance on the snow testing set of CSD, which is by $0.6$ dB higher than the corresponding performance of the snow-specific DRN model (as shown in Table~\ref{tab:baseline_drn_models}). When cross-stitch units are ablated from the above model hypothesis, the performance on the rain and snow data of the CSD dataset becomes worse; (b) the next four CMUDRN model hypotheses either switch cross-stitch units and alternate among using only the SSIM loss function or the Frobenius tensor norm loss function. Notably, we observe that when the SSIM loss function is used alone in a model hypothesis (ignoring the Frobenius tensor norm loss function), then the resulting model optimizes the SSIM performance of the resulting models both on the rain and snow data of CSD. However, when switching on cross-stitch units among these two model hypotheses, the model with multi-task learning behaves better than the one with no multi-task learning; and, (c) when using one of the two local loss functions in the CMUDRN model, the Frobenius tensor norm loss function appears to be associated with resulting models that have better performance than models trained only with the Frobenius tensor norm loss function and no cross-stitch units.

Finally, the last row on Table~\ref{tab:drn_model_ablation_study} is similar to the observed best model (in terms of the PSNR score), except that local loss functions are not applied on training the separate DRN submodules in CMUDRN. This is a baseline experiment to supplement the first CMUDRN model hypothesis case. We observe that this baseline model performs worse than the best model hypothesis (at the top row). Therefore, local supervision on the separate DRN submodules is important for training these task-specific submodules.

\section{Comparative experiments}
\label{sec:comparative_experiments}

\begin{table}[]
\caption{PSNR and SSIM scores generated for the CSD dataset by baseline Dual Recursive Networks. The first $70\%$ of the dataset examples amounting to $5600$ examples are used for training a baseline DRN model and the rest of the data are kept for model testing.}
\label{tab:baseline_drn_models}
\centering
\begin{tabular}{l|l|c|cc}
\textbf{experiment} & \textbf{dataset} & \textbf{iterations} & \textbf{PSNR} & \textbf{SSIM} \\ \hline
DRN@rain & CSD@rain     & 7, 7 &  28.35    &  0.91    \\ 
DRN@snow & CSD@snow     & 7, 7 &  23.02    &  0.89    \\ \hline
Chen et al. \cite{Chen_2022_CVPR} & CSD@snow & n/a & 31.33 & 0.94\\
HDCWNet \cite{chen2021all} & CSD@snow & n/a & 29.45 & 0.92\\
\end{tabular}
\end{table}

Our comparative experiments comprise two types of experiments: (a) baseline comparative experiments using the DRN model due to Cai et al. \cite{cai2019dual}; and, (b) third-party comparative experiments that are performed on the rain and snow CSD data that we reuse in this paper. Figure~\ref{fig:example_restored_images} shows example snowy and rainy images from the CSD dataset that are restored by the best CMUDRN model hypothesis from our ablation study in Section~\ref{sec:ablativestudy}. 

The baseline experiments are DRN models that are trained on rain or snow training data. They use $7$ intra and inter-iterations. The rain-specific DRN baseline model attains a PSNR value of $28.35$ dB and an SSIM value that equals $0.91$. The snow-specific DRN module attained a PSNR score of $23.02$ dB and an SSIM score of $0.89$.

The comparative experiments are conducted on the methods: (a) by Chen et al. \cite{Chen_2022_CVPR}, which is a recent state-of-the-art method; and, (b) the HDCWNet method by Chen et al. \cite{chen2021all}. The method by Chen et al. \cite{Chen_2022_CVPR} is a high performance SOTA model that scores a PSNR of $31.33$ dB on the snow testing data portion of the CSD dtaset. A lower PSNR performance is attained by the HDCWNet method, that equals $29.45$ dB. We observe that the SSIM values computed for these last two methods are $0.94$ and $0.92$.

Our experiments were implemented in the Python $3$ programming language using the PyTorch framework. We used the Adam optimization method \cite{kingma2014adam} and set the learning rate to $10^{-4}$. The experiments reported in this section were conducted on an NVIDIA GTX 1070 Ti GPU.

\section{Lightweight CMUDRN models}
\label{sec:performance_analysis}

In this section, we evaluate CMUDRN in terms of its empirical processing speed. We notice that there are two factors that contribute to the ability of the model to be fast: (a) the size of the input image; and, (b) the intra and inter-loop count. To measure the speed of the model in this setting, we generate equi-length random images of a size ranging from $100 \times 100$ to $1000 \times 1000$ pixels with a size step of $50$ pixels. For each image size, we generate $1000$ images with random pixels drawn from a normal distribution.

\begin{figure}
    \centering
    \includegraphics[scale=0.3]{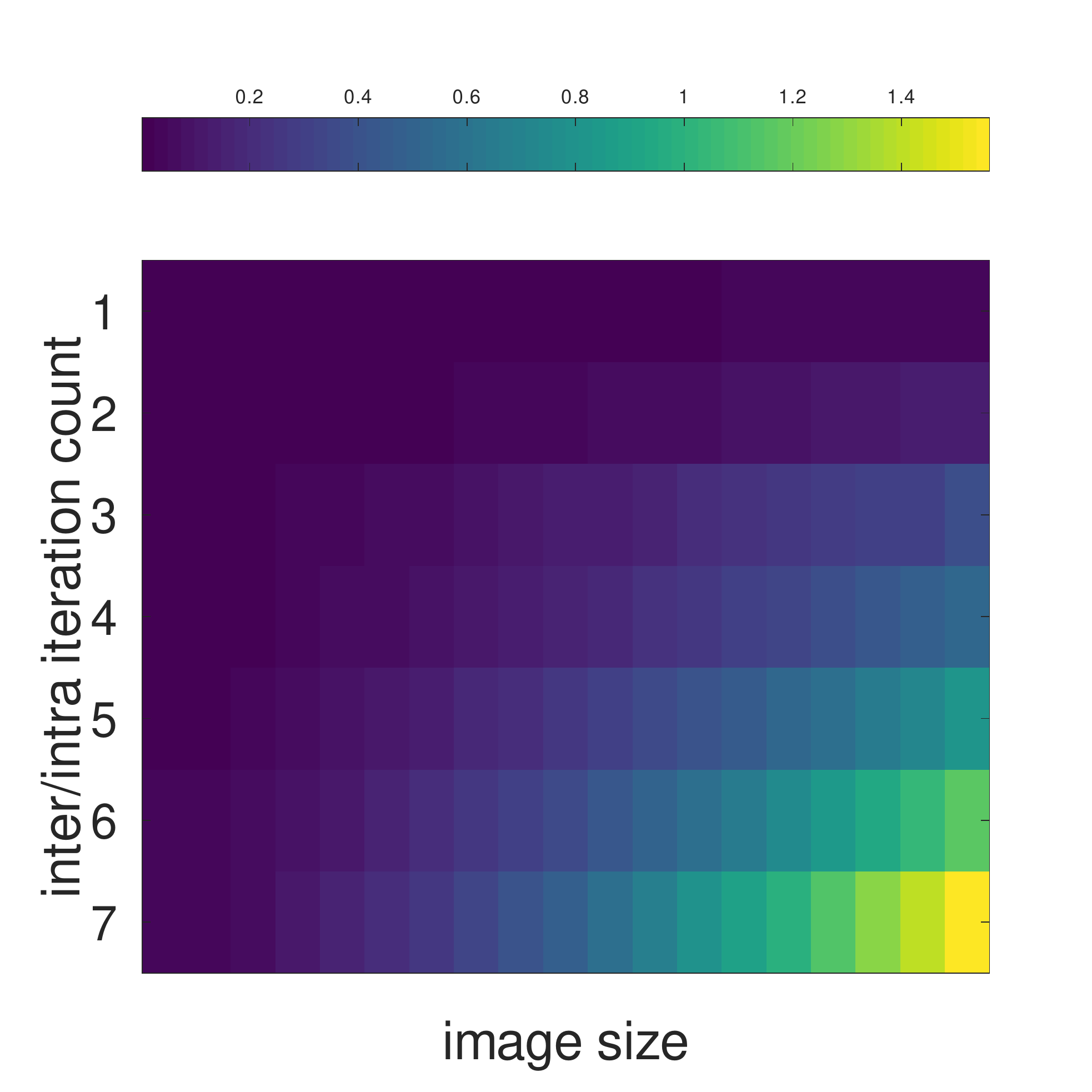}
    \caption{Heatmap of the average processing time (measured in seconds) consumed by the CMUDRN model when restoring a $n \times n$ image with randomly computed pixels. We vary the image border size $n$ from $100$ pixels to $1000$ pixels with a step of $50$ pixels. We vary the inter and intra-iteration count in the CMUDRN model from $1$ to $7$ iterations. Qualitatively, as we require a higher iteration count budget and increase the image size, the average running time requirement increases. A relatively fast CMUDRN model is obtained when a combination of an iteration count around the value $4$ is chosen and when the input image size is around $500 \times 500$ pixels. As these parameters grow, the average running time grows linearly. The maximum average running time is rendered for an iteration count equal to $7$ and an image size equal to $1000 \times 1000$ pixels.}
    \label{fig:processing_time_heatmap}
\end{figure}

To study the running time performance of the CMUDRN model, we leverage random image samples and the best CMUDRN model that we trained in our ablation study in Section~\ref{sec:ablativestudy}. Given this model, we vary the inter and intra-loop count and evaluate the model on random images of a varying size. Given parameters (a) and (b), we evaluate the model on $1000$ random images and then observe the running time of a forward pass in our model. We consider a loop count in the range from $1$ iteration to $7$ iterations, and an image size from $100 \times 100$ pixels to $1000 \times 1000$ pixels with a step of $50$ pixels. Figure~\ref{fig:processing_time_heatmap} illustrates the running time requirement of this CMUDRN model for a particular combination of loop count and input image size. Cai et al. \cite{cai2019dual} also studied the influece of the loop count parameter in the DRN model. Their numerical results suggest that changing the loop count does not impact the observed PSNR and SSIM metrics significantly, although this parameter determines the running time performance of the model.

The heatmap in Figure~\ref{fig:processing_time_heatmap} compactly represents the empirical running time of the CMUDRN model for a particular combination of parameters of type (a) and (b). We empirically observe that a low loop count and a small image size lead to a running time requirement that is low. For a small loop count, as the image size increases the running time requirement grows. Moreover, when the loop count budget and the image size grows, we finally observe a maximum runnning time requirement, that especially attains a maximum when the loop count approaches the value of $7$ and the input image size is $1000$ pixels.

Using a CMUDRN model with an iteration count equal to $3$ and an image size of $500 \times 500$ pixels, the model can afford 9 FPS. The same model with an input image of $1000 \times 1000$ pixels affords $2.55$ FPS. An optimized model with $7$ iterations and an input image size of $500 \times 500$ pixels affords $2.40$ FPS. Consequently, the same model affords $0.64$ FPS when the input image is of size $1000 \times 1000$ pixels.

\section{Conclusions}
\label{sec:conclusions}

We presented the Cross-stitched Multi-task Unified Dual Recursive Network (CMUDRN) model for unified deraining and desnowing that is reusing the architecture of the Dual Recursive Network (DRN) for learning task-specific models for rain and snow image data as a basic network module, whose outputs are fused together to allow for unified single image deraining and desnowing. We created a parametric neural bottleneck layer that combines the restored output of both DRN models and merges the contribution of both models into a unique restored image. We came up with a multi-task learning model (using cross-stitch units that allow for feature sharing among the DRN models) with a very low number of model parameters that at the same time provides good PSNR and SSIM performance scores on a controlled dataset of synthetic rainy and snowy images.

\section{Acknowledgements}
\label{sec:acknowledgements}

This research has been supported by the European Commission within the context of the project RESCUER, funded under EU H2020 Grant Agreement~101021836.

\bibliographystyle{IEEEtran}
\bibliography{bibliography}

\end{document}